\pdfoutput=1

\documentclass[11pt]{article}

\usepackage{emnlp2021}

\usepackage{times}
\usepackage{latexsym}
\usepackage{amssymb}
\usepackage{amsmath}
\usepackage{amsthm}
\usepackage{graphicx}
\usepackage{multirow}
\usepackage{booktabs}
\usepackage[nameinlink]{cleveref}

\usepackage[T1]{fontenc}

\usepackage[utf8]{inputenc}

\usepackage{microtype}

\title{The Low-Dimensional Linear Geometry \\ of Contextualized Word Representations}

\author{Evan Hernandez \\
  MIT CSAIL \\
  \texttt{dez@mit.edu} \\\And
  Jacob Andreas \\
  MIT CSAIL \\
  \texttt{jda@mit.edu} \\}

\begin{document}
\maketitle
\begin{abstract}
Black-box probing models can reliably extract linguistic features like tense,
number, and syntactic role from pretrained word representations. However, the
manner in which these features are \emph{encoded} in representations remains
poorly understood.
We present a systematic study of the linear geometry of contextualized word representations in ELMO and BERT.
We show that a variety of
linguistic features (including structured dependency relationships) are encoded in low-dimensional subspaces. We then refine this geometric picture, showing that there are hierarchical relations between the subspaces encoding general linguistic categories and more specific ones, and that low-dimensional feature encodings are distributed rather than aligned to individual neurons.
Finally, we demonstrate that these linear subspaces are causally related to model behavior, and can be used to perform fine-grained manipulation of BERT's output distribution.
\end{abstract}

\section{Introduction}

Contextual word representations \citep{ELMo}
encode general linguistic features (e.g.\ semantic class; \citeauthor{BelinkovSurvey}, \citeyear{BelinkovSurvey}) and sentence-specific relations (e.g.\ syntactic role; \citeauthor{TenneyBertRelearns}, \citeyear{TenneyBertRelearns}). Used as input features, pre-trained representations enable efficient training of models for a variety of NLP tasks \cite{peters2019tune}.
An enormous body of recent work in NLP has attempted to enumerate \emph{what}
features are encoded by pretrained word representations
\cite[e.g.][]{shi2016does,TenneyWhatDoYouLearn,BelinkovSurvey},
and more recent approaches have studied \emph{how accessible} this information is, characterizing the tradeoff between generic notions of model complexity and accuracy needed to recover word features from learned representations \citep{Voita,pimentel2020pareto}. But the manner in which these features are \emph{encoded} in representations remains poorly understood.

\begin{figure}[t!]
\vspace{-1em}
    \centering
    \includegraphics[width=\columnwidth]{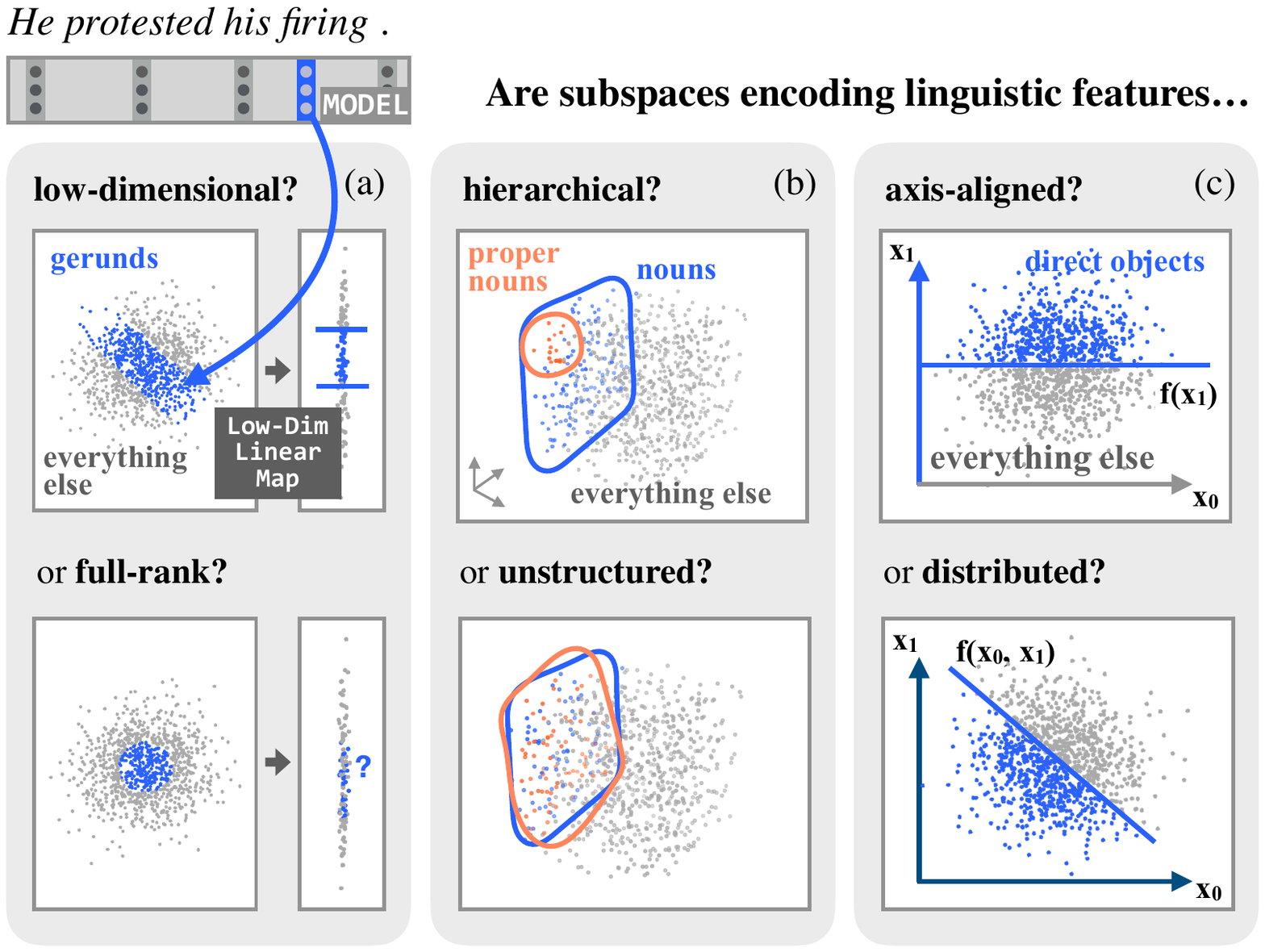} \\
    \vspace{-.75em}
    \includegraphics[width=\columnwidth,clip,trim=0 5.8in 0 0]{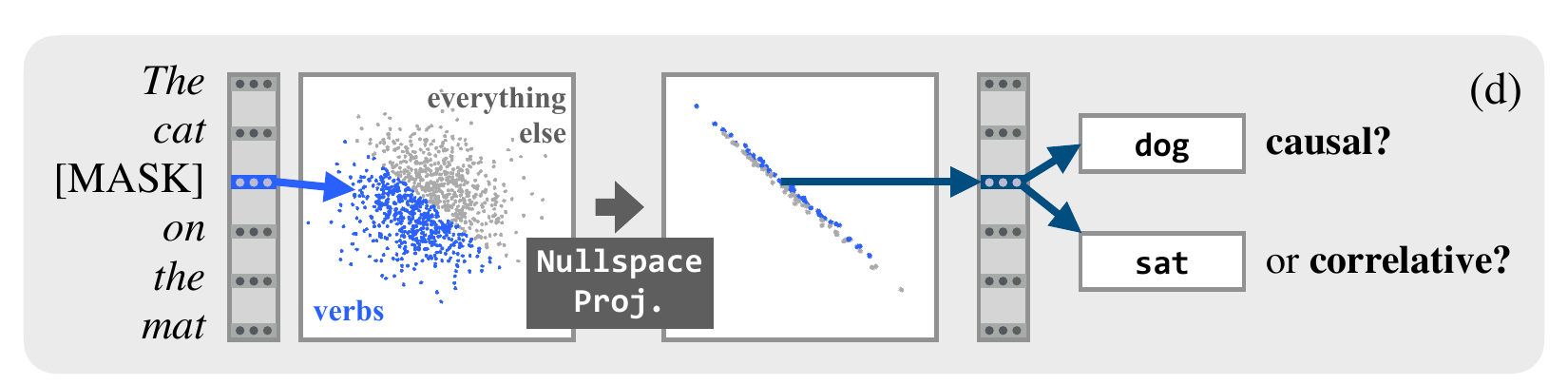} \\
    \caption{
      By training rank-constrained probing models on linguistic analysis tasks, we find that linguistically meaningful categories are (a) encoded in low-dimensional representation spaces; (b) organized hierarchically, but (c) not aligned with individual neurons. An additional set of experiments (d) uses these findings to identify linear operators that predictably affect outputs of a masked language model. Best viewed in color.
    }
    \label{fig:teaser}
    \vspace{-1em}
\end{figure}

In this paper, we investigate the \textbf{shape and structure} of learned representation spaces (\Cref{fig:teaser}).
We build on previous studies that have identified specific \textbf{linear subspaces} responsible for gender bias in word embeddings \cite{bolukbasi2016man} and word sense disambiguation \cite{coenen2019visualizing}.
We show that these linear subspaces are the rule, not the exception: a variety of other lexical and syntactic phenomena are encoded in low-dimensional linear subspaces, and these subspaces exhibit relational structure that has not yet been fully characterized by previous work.

Our approach involves training a sequence of expressive but \textbf{rank constrained} probing models, building on similar exploratory experiments by \citet{StructuralProbe} and \citet{ControlTasks} that study the effects of rank constraints on probe accuracy and complexity. We generalize the use of rank constraints to identify the lowest-dimensional subspace that encodes a task, finding that
(1) linguistic features specifically reside in \emph{low-dimensional} subspaces (\Cref{sec:sweep}) and (2) these subspaces exhibit some degree of \emph{hierarchical structure} (\Cref{sec:hierarchy}), but (3) are mostly \emph{not aligned} with individual neurons (\Cref{sec:axis-alignment}). 
Additional findings include that linguistic features tend to be encoded in lower-dimensional subspaces in early layers of both ELMO and BERT, and that relational features (like dependency relations between pairs of words) are encoded less compactly than categorical features like part of speech.

As an example of how this kind of information can inform ongoing NLP research, we conclude with a demonstration that discovered subspaces can be used to exert fine-grained control over masked language models: specific linear transformations of the last layer of BERT narrowly ablate its ability to model agreement in nouns and verbs (\Cref{sec:ablation}).

\section{Related Work}

The predominant paradigm for interpreting learned word representations is 
based on training auxiliary \emph{probing} models to predict linguistic attributes from fixed representations. For standard pretraining schemes, simple probes can successfully recover parts of speech, dependency relations, semantic role labels, coreference links, and named entities \cite{TenneyBertRelearns, TenneyWhatDoYouLearn, Liu}. These features are better encoded by language models than encoders trained for machine translation \cite{TranslationProbing}.
More complex probes target structured labels such as dependency trees \cite{StructuralProbe} and parser state \cite{NLMSyntacticState}.
Other probes localize linguistic information further: in time, as a function of training dynamics \cite{SaphraLopez},
and in space, e.g.\ as a function of individual hidden units \cite{BelinkovNeurons, GrainOfSand}. 
See  \citet{BelinkovSurvey} for a detailed survey.

Our work falls into the latter category: we aim to identify \emph{global,
linear} structure in representation spaces. Some previous work has identified surprising non-linear
geometry in word embeddings \citep{mimno2017strange}, but other work suggests that simple linear
subspaces encode meaningful attributes like gender information \citep{bolukbasi2016man,
vargas2020exploring, INLP} and word sense information
\citep{coenen2019visualizing,ethayarajh2019contextual}.
We extend this account to a variety of other linguistic
features, identify relations among feature subspaces themselves, and show
that these subspaces are causally linked to model behavior.

More recent work has focused on addressing shortcomings of the probing paradigm itself. \citet{ControlTasks} design non-linguistic control tasks to benchmark how selective probes are for linguistic information, and show that high probe accuracy is sometimes attributable to task simplicity rather than explicit representation of linguistic content.
\citet{VoitaTransformers}, \citet{Voita} and \citet{Pimentel} describe an information-theoretic approach to probing. \citet{Voita} in particular argue that measurements of probe quality based on \emph{description length} characterize encodings more precisely than raw selectivity measurements,
showing that part-of-speech tags and dependency edges can be extracted with a smaller code length than controls.
\citet{ProbingProbing} and \citet{elazar2021amnesic} further question the use of probe accuracy, finding that language models encode linguistic information even when it is not causally linked to task performance.
In contrast, this paper aims to characterize what information is encoded \emph{in any form} in low-dimensional representation subspaces, regardless of the complexity of the encoding.
\section{Method}
\label{sec:method}

Consider a corpus of words $W = (w_1, \dots, w_n)$ with $D$-dimensional word representations $(r_1, \dots, r_n)$ written as a matrix $R \in \mathbb{R}^{D \times n}$. A \textbf{probing task} on $W$ is defined by a mapping $f(W)$ from words to discrete labels (part of speech tags, head--dependent relations, semantic roles, etc.) We say the representations \textbf{encode} the task if there exists a \textbf{probe} $g(R)$ that predicts $f(W)$ according to a \textbf{score} $\mathcal{S}(g(R), f(W))$, which we choose to be \emph{held-out accuracy}.
There are other choices for $\mathcal{S}$, e.g.\ selectivity \cite{ControlTasks} and MDL \cite{Voita}, but these measure different aspects of probe complexity, whereas we are interested in the existence of \emph{any information} in subspaces of $R$ that is predictive of $f(W)$.

We aim to find the \textbf{lowest-dimensional subspace} $\mathbb{R}^d$ of $\mathbb{R}^D$ within which the representations $R$ encode the linguistic task $f(W)$. %
If $\Pi$ is a projection from $\mathbb{R}^D$ to the lower-dimensional
$\mathbb{R}^d$, we say the subspace encodes the task if the optimal probe
predicting task labels from $\Pi R$ is approximately as accurate as the optimal probe predicting them from $R$.
Formally, given a tolerance $\alpha$, we seek the smallest positive integer $d <
D$ for which there exists $\Pi_d \in \mathbb{R}^{d \times D}$ and probes $g$ and $g'$ satisfying:
\begin{equation}
\mathcal{S}\big(g(R), f(W)\big)
-
\mathcal{S}\big(g'(\Pi_d R), f(W)\big)
\le \alpha
\label{eq:probing}
\end{equation}

In practice, we cannot optimize directly for $d$ via gradient descent because matrix rank is neither convex nor differentiable. Instead, we simply enumerate values of $d$ and, for each rank, learn a $d$-dimensional projection\footnote{
    ``Projection" is a slight abuse of terminology: we do not guarantee that
    $\Pi^2 = \Pi$. For the purposes of our experiments, the rank constraint on
    $\Pi$ is sufficient.
}
jointly with a probe $g$ implemented as a multilayer perceptron.
We use the same architecture for all tasks, beginning with an investigation of the value of $d$ in a set of standard probing experiments.
The use of an expressive model \emph{downstream} of the projected representation disentangles questions about \emph{representation geometry} from the questions about \emph{probe capacity} considered in previous work.\footnote{Supplementary experiments, which show slightly worse results for linear probes,
indicate that feature encodings are indeed low-dimensional but nonlinear (see \Cref{app:linear-probes}).}

\section{Are linguistic variables encoded in low-dimensional subspaces?}
\label{sec:sweep}

Our first set of experiments aims to characterize the subspaces that encode
linguistic tasks in contextualized representations. Following \citet{ControlTasks}, we experiment with two contextualized representations alongside their non-contextual word embeddings, and we probe them on three linguistic tasks and three non-linguistic control tasks.

\subsection{Tasks}

Our three tasks explore a set of core syntactic phenomena, including syntactic categories, roles, and relations. These tasks are used as a standard testing battery in the probing literature \cite{ControlTasks, Voita, TenneyBertRelearns, TenneyWhatDoYouLearn}, and prior work on full-rank representations has found high probing accuracy for all three.

\textbf{Part of Speech Tagging (POS)} is the task of predicting one of 45 part of speech tags (e.g. ``plural noun'' / NNS) for a word in context.

\textbf{Dependency Label Prediction (DLP)} is the task of predicting one of 45 syntactic relationships between a pair of words connected by a dependency edge in a ground-truth parse tree.

\textbf{Dependency Edge Prediction (DEP)} is the more challenging task of predicting dependency edges themselves: given a word, we try to predict the index of its head.

For all tasks, we use the Penn Treebank \cite{PTB} with the standard train, development, and test splits to train and evaluate probes.
Like previous work, we compare results with control tasks, which feature the same label structure but arbitrary mappings from inputs to labels. These controls are the same as in \citet{ControlTasks} for POS and DEP; for DLP, we map pairs of word types to one of 45 tags at random.

\subsection{Representations}

We probe two sets of word representations.

\textbf{ELMo} \cite{ELMo} is a 2-layer bidirectional LSTM trained for language modeling. We use the 5.5 billion-word pre-trained model and treat both 1024-dimensional hidden states as separate representations (layers 1 and 2). We also include the non-contextual word embeddings produced by the character CNN (layer 0). 

\textbf{BERT} \cite{BERT} is a transformer trained on masked language modeling. We use the pre-trained base version from \citet{HuggingFace}. It has 12 attention heads, each producing 768-dimensional hidden representations. We treat the output of four different heads (layers 1, 4, 8, and 12) as separate representations and also include the initial word embedding layer (layer 0).\footnote{We include results for untrained BERT in \Cref{sec:random}.}

\subsection{Probes and optimization}

\begin{figure*}
    \centering
    \includegraphics[width=\textwidth]{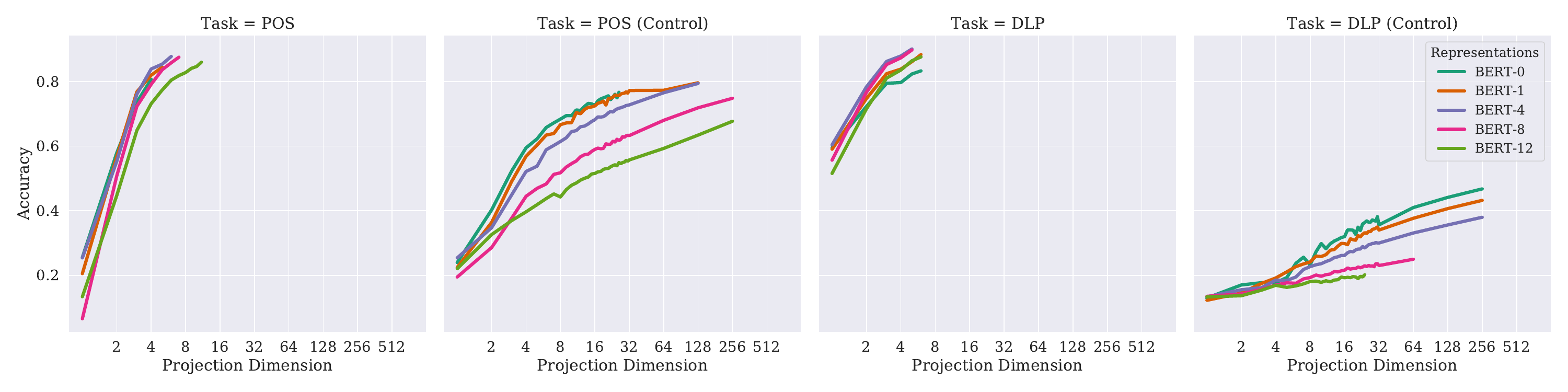}
    \vspace{-2em}
    \caption{
        Accuracy as a function of projection rank grouped by task, representation model, and representation layer. Each line terminates when it achieves accuracy within $\alpha=0.05$ of the best achievable accuracy for that layer; lines ending closer to the left side of the graph indicate lower-dimensional representations.
    }
    \label{fig:acc_vs_dim}
\end{figure*}

\begin{table*}[t]
    \centering
    \footnotesize
    \begin{tabular}{ccllllll}
    \toprule
    \multicolumn{2}{c}{\multirow{2}{*}{}} & \multicolumn{2}{c}{\textbf{POS}} & \multicolumn{2}{c}{\textbf{DLP}}  & \multicolumn{2}{c}{\textbf{DEP}} \\ %
    \multicolumn{2}{c}{} & \multicolumn{1}{c}{Real} & \multicolumn{1}{c}{Control} & \multicolumn{1}{c}{Real} & \multicolumn{1}{c}{Control} & \multicolumn{1}{c}{Real} & \multicolumn{1}{c}{Control} \\ \midrule %
    \multicolumn{1}{c}{\multirow{3}{*}{\textbf{ELMo}}} & 0  & .92 / 6 & .95 / 256 & .86 / 5 & .55 / 512 & .37 / 11 & .76 / 13 \\
    \multicolumn{1}{c}{} & 1  & .93 / 5  & .88 / 256 & .93 / 3 & .40 / 512 & .84 / 13 & .82 / 17 \\ 
    \multicolumn{1}{c}{} & 2  & .93 / 6 & .82 / 256 & .92 / 4 & .33 / 256 & .79 / 21  & .78 / 23 \\ \midrule %
    \multicolumn{1}{c}{\multirow{5}{*}{\textbf{BERT}}} & 0  & .81 / 4 & .77 / 26 & .83 / 6 & .47 / 256 & .60 / 13 & .80 / 7 \\
    \multicolumn{1}{c}{} & 1  & .84 / 5 & .80 / 128 & .88 / 6  & .43 / 256 & .71 / 17 & .80 / 9 \\ 
    \multicolumn{1}{c}{} & 4  & .88 / 6 & .80 / 128 & .90 / 5 & .38 / 256 & .77 / 13 & .79 / 10 \\ 
    \multicolumn{1}{c}{} & 8  & .88 / 7 & .75 / 256 & .90 / 5 & .25 / 64 & .81 / 13 & .71 / 12 \\ 
    \multicolumn{1}{c}{} & 12 & .85 / 10 & .68 / 256 & .88 / 6 & .20 / 24 & .76 / 20 & .65 / 11 \\ \bottomrule %
    \end{tabular}
    \caption{
     Each table entry reports two numbers: the \textbf{accuracy} / \textbf{dimensionality} an approximately optimal $d$-dimenional probe (Equation~\ref{eq:probing}, with $\alpha=0.05$).
     In general, real tasks are encoded in lower-dimensional subspaces than control tasks; in BERT, representations become more diffuse (distributed across more dimensions) in deeper layers. \Cref{fig:acc_vs_dim} shows the full accuracy-vs-dimension tradeoff curve for ELMO on the POS and DLP tasks.
    }
    \label{tab:acc-vs-dim}
\end{table*}

\textbf{Method.} For each task and representation, we apply the method described in \Cref{sec:method} to find the optimal $d$-dimensional subspace encoding that feature. We sweep over all smaller dimensions $d=1,2,\dots,32$ and exponentially over the larger $d=64, 128, \dots, D$. At each iteration, we train an MLP to predict the feature given the projected representations as input.
While \Cref{eq:probing} is technically an expectation, preliminary experiments revealed that $d$ is relatively insensitive to random restarts, so we train one probe per configuration.

\textbf{Architecture.} We use a 2-layer MLP of the form $\textsc{MLP}(x) = \textsc{softmax}(W_2\textsc{ReLU}(W_1x))$ for all experiments. The hidden layer always has the same size as the input layer, and only the interpretation of the input and output varies. For POS, $x$ is a single word representation and $\textsc{MLP}(x)$ is a distribution over POS tags for that word. In DLP, $x$ is the concatenation of two representations $x = [h_i;  h_j]$ and $\textsc{MLP}(x)$ is a distribution over dependency labels for word $i$ and $j$. For DEP, $x = [h_i; h_j]$ is the same as in DLP and $\textsc{MLP}(x)$ is the probability that word $i$ is the parent of word $j$ in a given sentence.

\textbf{Optimization.} We minimize the negative log-likelihood of the probe on the task dataset using Adam \cite{Adam} with a learning rate of 0.001, and we stop training when the loss on the development set does not improve for 4 epochs or when we exceed 1000 epochs. 

\subsection{Results}

\Cref{tab:acc-vs-dim} shows the highest accuracy of any probe on each task for each representation model and layer, and \Cref{fig:acc_vs_dim} plots accuracy against projection rank for all BERT layers on POS and DLP. From these, we can draw several conclusions.

\textbf{Linguistic variables are encoded in low-dimensional subspaces.} For POS and DLP across all representations, the probe requires only a $d < 10$ dimensional subspace to reach within $\alpha = 0.05$ of its optimal accuracy. DEP appears to be more challenging, but still performs at optimality within a $d < 21$ dimensional subspace. We see that layer 1 of ELMo and early layers of BERT solve real tasks with a lower-dimensional subspace than the other layers. These results agree with \citet{Voita}, who find that the middle layers of ELMo and BERT can be used to train probes of shorter description length.\footnote{Note that these results cannot be explained by the \emph{anisotropy} of word representations, a phenomenon previously observed by \citet{mimno2017strange} and \citet{ethayarajh2019contextual}---the comparatively high-dimensional subspaces for control tasks indicate that low dimensionality is a specific property of linguistic features, not of embeddings themselves.}

\textbf{Contextual representations encode linguistic variables better, but not more compactly, than non-contextual word embeddings.} 
Probes trained on the contextual representations outperform those trained on the word embeddings, often by a substantial margin, but sometimes with an equal or greater number of subspace dimensions. For example, a probe trained on BERT-0 achieves 81\% accuracy on the POS task with only a 4-dimensional subspace, while BERT-8 can achieve 88\% accuracy at the task, but requires almost twice as many dimensions.
Like previous work, we find that contextualization is 
important for good probe performance, but that probe-relevant information is sometimes more distributed in deeper layers.

\textbf{Non-linguistic control tasks are not encoded in low-dimensional subspaces.} 
Probes generally struggle to learn non-linguistic control tasks from the projected representations, with the worst probes being trained on BERT-12 and achieving 68\% accuracy on the POS control and 20\% accuracy on the DLP control. Our results coincide with those of \citet{ControlTasks}, who argue that word embeddings outperform contextual representations on control tasks because they better encode word identity, which is the sole factor of variation in the control task.

One exception to our findings is the DEP control. We conjecture simply that it is too easy. The task has only three unique labels, so the
MLP might easily find a nonlinear boundary that memorizes the label for each word. This is corroborated by the weaker accuracies of linear probes trained on the DEP control shown in 
\Cref{app:linear-probes}.

\section{How are these subspaces structured?}

Next,
we identify two important geometric properties of these low-dimensional subspaces.

\subsection{Hierarchy}
\label{sec:hierarchy}

\begin{table}
\footnotesize
\centering
\begin{tabular}{ll}
\toprule
Task & Tags \\
\midrule
POS-Noun & NNP, NNPS, NN, NNS \\
POS-Noun-Proper & NNP, NNPS \\
POS-Verb & VBP, VBG, VBZ, VB, VBD, VBN \\
POS-Verb-Present & VBP, VBG, VBZ \\
\bottomrule
\end{tabular}
\caption{\label{tab:hierarchy} Label sets for hierarchical POS subtasks. All other labels are collapsed into an N/A tag.}
\vspace{-1em}
\end{table}

Part of speech tagging is typically treated as a categorical problem, but in
reality the labels have more structure: nouns (distributed across tags
NN, NNS, etc.) behave more like each other than verbs. Our next question is whether similar structure manifests inside subspaces of word representations. Specifically, within the subspace encoding part of speech, is there a still lower-dimensional subspace that suffices for distinguishing nouns from each other? And within that subspace, is there yet a lower-dimensional sufficient for labeling proper nouns? We find two examples of such subspace hierarchies.

\begin{figure}[t]
    \centering
    \vspace{-.5em}
    \includegraphics[width=.52\textwidth]{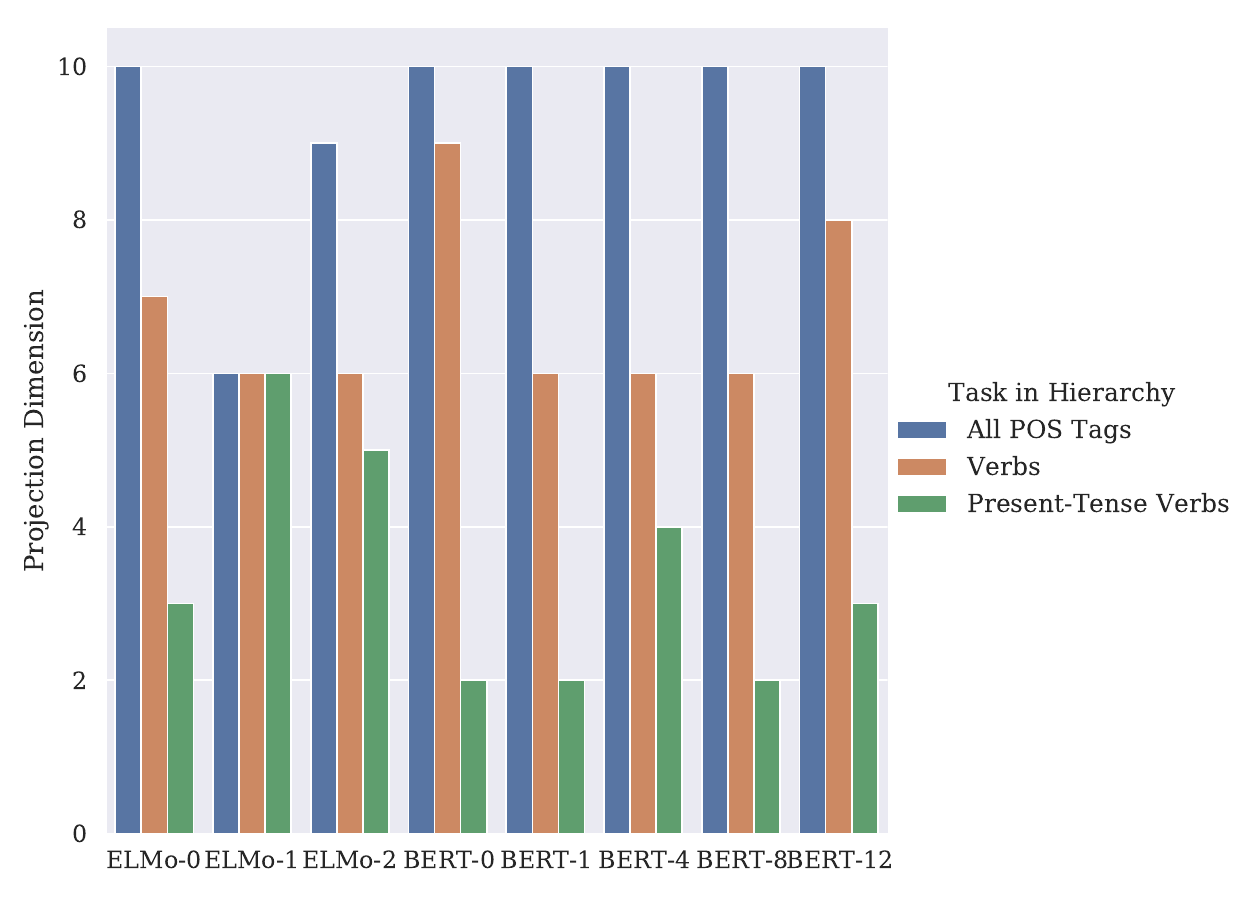}
    \includegraphics[width=.47\textwidth]{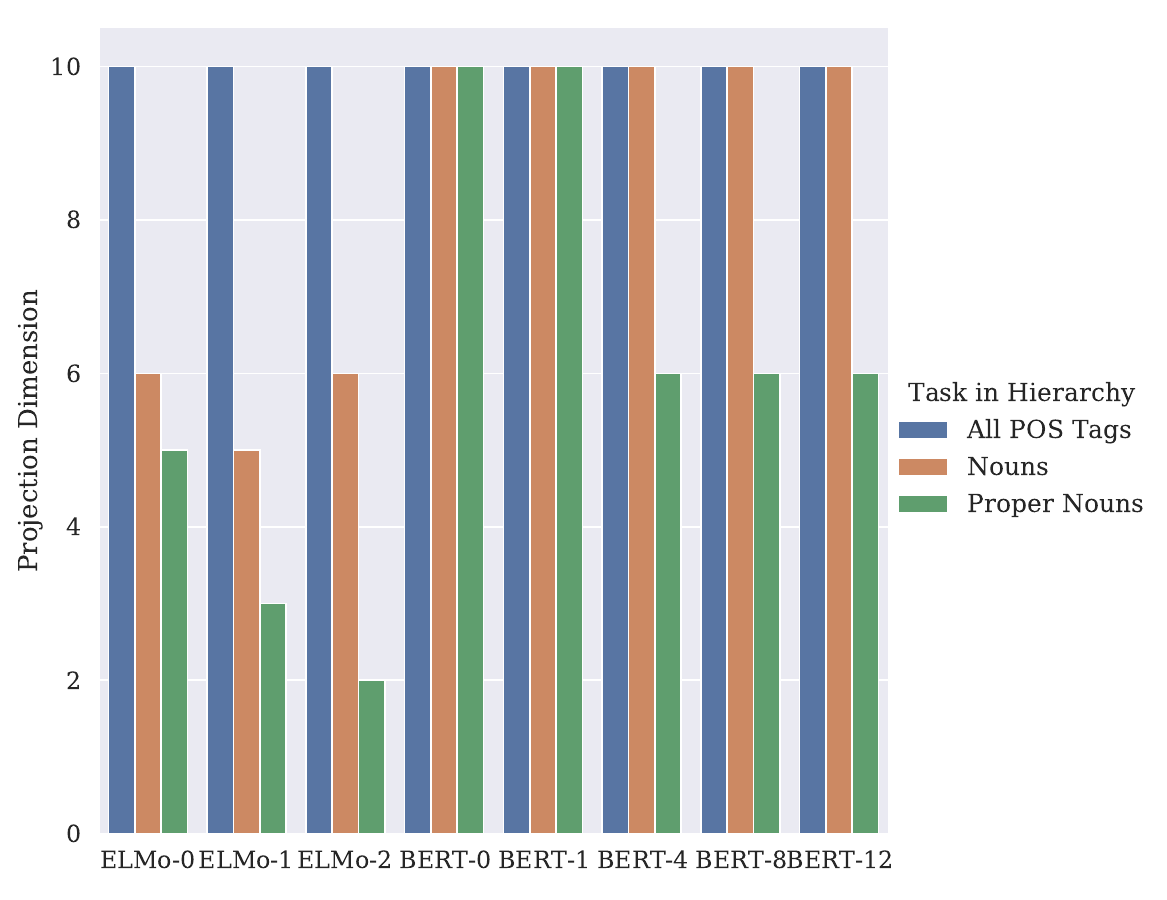}
    \vspace{-.5em}
    \caption{
        Lowest-dimensional projection for each task in the hierarchy, grouped by representation and ordered by the number of tags in the task. An orange (green) bar shorter than its blue (orange) neighbor represents a low-dimensional subspace encoding of a task (e.g. POS-Verb-Present) that lives \emph{inside} of another subspace encoding a larger task (e.g. POS-Verb). Many representations appear to express these hierarchies.
    }
    \label{fig:hierarchies}
    \vspace{-1em}
\end{figure}

\begin{figure*}
    \centering
    \includegraphics[width=.90\textwidth]{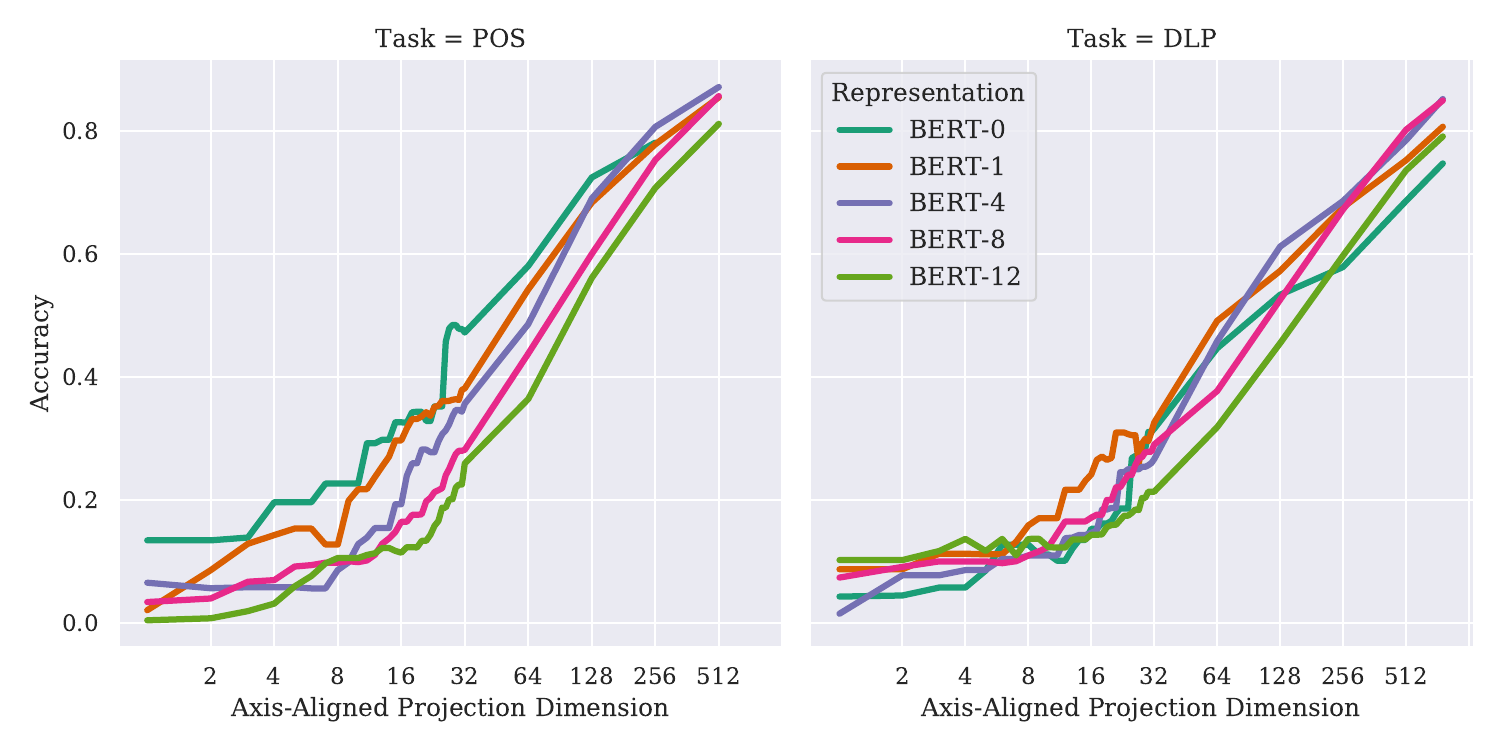}
    \vspace{-1em}
    \caption{
        Test accuracy as a function of number of \textbf{nonzero axes/neurons}. As in \Cref{fig:acc_vs_dim}, lines terminate once a $d$-dimensional probe achieves accuracy within $\alpha = 0.05$ of the optimum. The optimal axis-aligned projections use more than half the representation axes: substantially higher than the ranks of the projections found in \Cref{sec:sweep}.
    }
    \label{fig:axis-alignment}
    \vspace{-1em}
\end{figure*}

\textbf{Method.} We first decompose the POS task into smaller tasks by selecting a set of related tags to keep and replacing the rest with a single ``N/A" tag. For example, in the POS-Verb task, we keep all tags related to verbs (VB, VBD, VBN, etc.) and re-tag any word that is not a verb with ``N/A''. \Cref{tab:hierarchy} shows the label sets we use in our experiments. 

For each task in a hierarchy, we apply our method from \Cref{sec:method} and sweep over projection ranks $d_1 = 1,2,\dots,d_0$ for a fixed $d_0$ to identify the lowest dimensional subspace in which the MLP probe achieves accuracy at least $\beta$ on the task, where $\beta$ is a fixed threshold. We then project all representations onto that subspace and proceed to the next task, training probes on the \emph{projected} representations instead of the full representations and sweeping over $d_2 = 1,2,\dots,d_1$. We set $\beta = .95$ and $d_0 = 10$ for all experiments.\footnote{
Early experiments showed that the MLP probe reaches near perfect accuracy when trained on the new POS tasks. Hence, the choice of $\beta = .95$ is equivalent to choosing $\alpha = .05$ in the framework of \Cref{sec:method}.
While the probe cannot always reach this accuracy for the \emph{full} POS task, we know from \Cref{sec:sweep} that the optimal $d$ with $\alpha = .05$ across any representation is $\le 10$, which is captured by our choice of $d_0$.
}

\textbf{Results.} \Cref{fig:hierarchies} plots for each task and each hierarchy the lowest $d$ for which the MLP probe obtains optimal accuracy in a $d$-dimensional subspace. We see that many, but not all, layers of the representations admit hierarchies of subspaces encoding the POS subtasks. BERT layers 0, 1, and 4 in particular do not appear to contain low-dimensional subspaces of their POS subspace that solve the noun task. In layers where hierarchies do manifest, the subspaces that solve the noun and verb subtasks are roughly half the rank of the POS subspace from which they were projected, suggesting that the fine-grained POS information is compactly encoded in the larger subspace that encodes all POS.

\subsection{Axis Alignment}
\label{sec:axis-alignment}

While linguistic information may be encoded in a low-dimensional subspace
of $\mathbb{R}^d$, there is no guarantee that any \emph{basis} for this
subspace bears any relationship to the neural network that produces representations $R$.
One refinement of the question from \Cref{sec:sweep} is whether
these subspaces are encoded by a small subset of neurons. Prior work finds neuron-level locality in a variety of models. \citet{BelinkovNeurons} find that machine translation models rely on a small subset of neurons, and \citet{SentimentNeuron} find a single recurrent unit that solves sentiment analysis. \citet{NetDissect} propose a general framework for probing units in convolutional neural networks, and \citet{GrainOfSand} similarly analyze the neurons of NLM and NMT networks. Both studies find that many neurons individually correlate with interpretable concepts.

We now consider a version of the experiment of \Cref{sec:sweep} restricted to \textbf{axis-aligned} projections constructed by selecting a subset of \emph{neuron activations}, i.e.\ rows of the representation matrix $R$.

\textbf{Method.} We first train a 10-dimensional probe and projection for POS and DLP, noting that from \Cref{sec:sweep} we know this probe will achieve within $\alpha = 0.05$ of the optimal accuracy for both tasks. We then zero each row of the projection, one at a time, and compute the probe's accuracy on the development set without tuning the probe or projection. This is equivalent to zeroing components of the input representation. After repeating this procedure for all rows, we record the index of the row that least reduced development-set accuracy when ablated, and permanently zero it. We then compute the accuracy on the test dataset, and repeat the full algorithm until all rows are zeroed.

\textbf{Results.} \Cref{fig:axis-alignment} plots test accuracy against axis-aligned projection size for BERT. As in \Cref{fig:acc_vs_dim}, lines terminating to the left of $x=768$ indicate the existence of accuracy-preserving subspaces. For POS, all layers can be reduced to 512 dimensions without a significant loss in accuracy, while for DLP some degradation begins immediately. However, aside from BERT-8 and BERT-12, all representations maintain over 70\% accuracy on POS and DLP until the last third of representation components are zeroed. In general, later BERT layers like BERT-8 and BERT-12 appear less axis aligned because probe accuracy decreases faster than it does other layers. A possible explanation is that low-level syntactic information like POS and DLP becomes more distributed as it is processed.

These results suggest that low-dimensional subspaces found in \Cref{sec:sweep} are indeed local to a subset of neurons, but the number of neurons may be much larger than the dimension of the subspace they support. It is important to note that, because of the greedy ablation procedure used in this section, we cannot rule out the existence of other, smaller sets of neurons providing the same projection accuracy; the results in \Cref{fig:axis-alignment} are an upper bound.

\section{Are low-dimensional encodings relevant to model behavior?}
\label{sec:ablation}

One critique of the probing paradigm is that probes do not reveal whether the model that produced the representations relies upon that information \cite{ProbingProbing}. Indeed, \citeauthor{ProbingProbing} construct datasets that do not require specific linguistic features and find that the natural language inference probes of \citet{Conneau} trained on their datasets detect those features nonetheless. How can we know that the low-dimensional linguistic features we uncovered in \Cref{sec:sweep} are used by the language model that constructed them?

Our geometric approach to probing lends itself to a simple interventional study. After computing the lowest-dimensional subspace encoding a task, we can remove representations from that subspace by projecting onto its nullspace. This should preserve all information in the representation except for the information needed for the target task.

The experiments that follow provide a small demonstration of this intervention.
We project representations from the final layer of BERT out of the subspace that encodes part of speech information. We then feed the ablated representations to a pre-trained language model with no fine-tuning and measure its performance on a challenge task that requires knowledge of part of speech distinctions.

\subsection{Task}

We use three groups of sentences from the dataset of \citet{marvin-linzen-2018-targeted} designed to test BERT's predictions at subject--verb agreement. Each suite consists of 13 sentences with four variants, one for each combination of singular/plural subject/verb (\Cref{tab:ablation}). All sentences contain a distractor phrase between the subject and the verb, and the distractor noun never agrees in tense with the subject noun. For example: \emph{The \underline{author} that the \underline{senators} hurt is okay}. The three test suites each use different distractors: one uses subject relative clauses, another uses prepositional phrases, and the final uses object relative clauses. We obtained the data from Syntax Gym \cite{SyntaxGym}.

This dataset lends itself to our interventional study because the sentences are highly structured and are designed to challenge language models. Moreover, satisfying subject-verb agreement in the presence of distractors requires fine-grained knowledge of part of speech. If BERT relies on the low-dimensional POS information encoded in its representation, then ablating the POS subspaces should impair its performance on the task.

\subsection{Ablation Method}
\label{ssec:ablation-method}

Our goal is to remove the low-dimensional linear part of speech information from BERT representations. Instead of removing all linear part of speech information, however, we will remove only subspaces that distinguish noun and verb parts of speech. We refer to these subspaces as \textbf{nounspace} and \textbf{verbspace}, respectively. Nulling out nounspace (verbspace) should damage BERT's ability to distinguish nouns (verbs) from each other.

We first apply our method from \Cref{sec:method} to find the lowest-dimensional subspace encoding the POS-Verb and POS-Noun tasks of \Cref{tab:hierarchy}. These subspaces have dimension 3 and 4, respectively. We then compute a projection onto their nullspaces. Letting $\Pi$ be the learned linear transformation and $U$ its left singular vectors, the projection onto the nullspace of $\Pi$ is given by:
\begin{equation}
N = I - UU^\top
\end{equation}
Note that our method is similar to, but distinct from, the INLP method of \citet{INLP}, which has previously been used to study the causal relationships between linear encodings of linguistic variables and BERT's predictions \cite{elazar2021amnesic}. INLP removes from the representations \emph{all subspaces} that are \emph{linearly predictive} of a set of labels. By contrast, our method ablates \emph{a single subspace} in which an \emph{arbitrary probe} can learn to predict a set of labels, allowing us to evaluate how BERT uses these subspaces to make predictions.\footnote{For comparison, we repeat our experiments using INLP in \Cref{app:inlp} and observe a less controlled effect.}

\subsection{Evaluation}

Let $\overline{N}$ and $\overline{V}$ be the nullspace projections for noun and verb information, resepectively.
For each sentence $s = (w_1, \dots, w_n)$ in each test suite, we mask either the subject noun or matrix verb to create the sentence $s_{masked}$ (e.g.\ constructing \emph{The \textrm{[MASK]} that the senators hurt is okay)}. We then feed $s_{masked}$
to the BERT transformer, recording only the output of the final layer.

Suppose we have masked the matrix verb. Let $r_v$ be the contextual representation of the mask token. We compute the original word probabilities $\textsc{MLM}(r_v)$, nounspace-ablated probabilities $\textsc{MLM}(\overline{N}r_v)$, and verbspace-ablated probabilities $\textsc{MLM}(\overline{V}r_v)$ for the verb slot. We then repeat this process with the subject noun masked instead of the verb, and record MLM output for the noun slot.

In both the normal and ablated decodings, we measure the difference in probability between the correct and incorrect word forms for the subject and matrix verb slots. We also measure the total probability mass that BERT assigns to any noun in subject slot and any verb in the verb slot.

\subsection{Results}
\label{ssec:ablation-results}

\begin{figure}[t]
    \vspace{-0.5em}
    \centering
    \includegraphics[width=.5\textwidth]{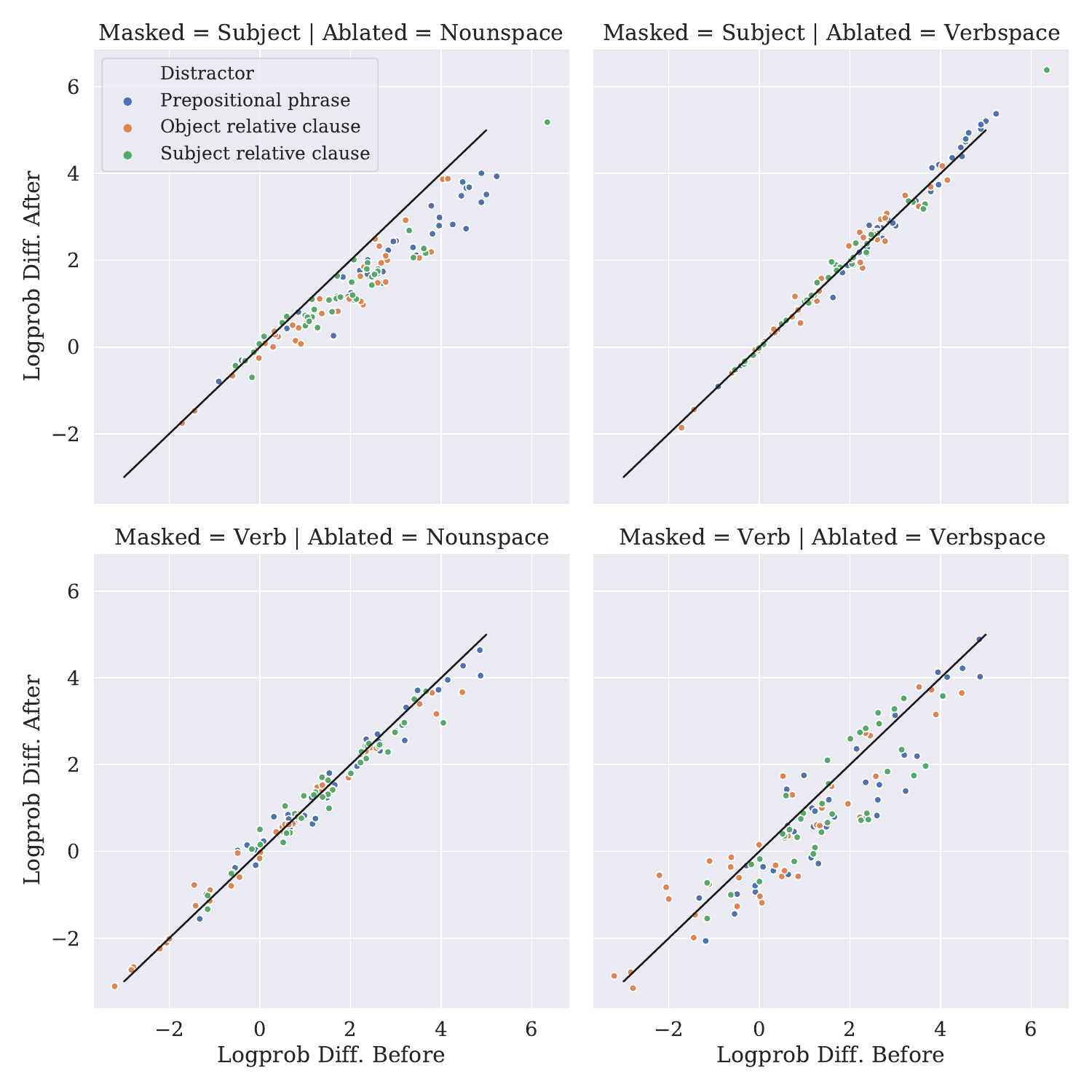}
    \caption{
        Difference in log probability of agreeing and disagreeing subjects (top) and verbs (bottom) before ($x$-axis) vs. after ($y$-axis) ablation of nounspace (left) and verbspace (right). For example, one point in the top left plots represents the difference in log probability between ``author'' and ``authors'' in ``The [MASK] that hurt the senators is good.'' Black line represents no change. Ablating nounspace increases BERT's confusion about the subject (points below the line), but has limited effect on its verb predictions (points on the line). Ablating verbspace has the opposite effect.
    }
    \label{fig:diffs}
    \vspace{-1em}
\end{figure}

\begin{table}[t]
    \centering
    \footnotesize
    \begin{tabular}{lccc}
    \toprule
    \multirow{2}{*}{\begin{tabular}[c]{@{}l@{}}Marginal\\Probability\end{tabular}} & %
    & Ablated &
    \\ \cmidrule{2-4}
    & Nothing & Verbspace & Nounspace \\ \cmidrule{1-4}
    Subject is noun & .85  & .85 & .82 \\ \cmidrule{1-4}
    Matrix is verb & .54  & .50 & .52  \\ \bottomrule
    \end{tabular}
    \caption{
        Probability mass assigned by BERT to nouns in the masked subject slot and verbs in the masked verb slot before and after ablation.
    }
    \label{tab:ablation}
    \vspace{-1em}
\end{table}

\Cref{fig:diffs} highlights that \textbf{both ablations substantially reduce BERT's performance on subject--verb agreement.} The difference in probability between the correct and incorrect subject (verb) forms consistently decreases after nullifying nounspace (verbspace). \textbf{This effect is highly controlled}: ablating nounspace does not appear to change BERT's ability to choose a verb that agrees with the subject, nor does ablating verbspace impair BERT in choosing a subject that agrees with the verb.

\textbf{Neither ablation substantially alters BERT's ability to distinguish nouns and verbs from other parts of speech}. \Cref{tab:ablation} shows that the verbspace ablation only slightly decreases the probability of BERT predicting a verb in the matrix verb slot, and does not at all decrease the probability of predicting a noun in the subject slot. We conclude that our \textbf{ablations are selective}: they \emph{only} impair BERT's ability to distinguish between subcategories of nouns and verbs, not its ability to reason about coarse parts of speech.

It is surprising that our ablations produce such fine-grained changes to BERT's outputs despite removing so little information from the representations. Both nullspace projections remove fewer than 1\% of the dimensions from one layer of the hidden representations for a single masked word. This justifies our interpretation of low-dimensional subspaces as minimal information-encoding units.

\section{Conclusions}

We have described a procedure for probing the linear geometry of word representations. Our method identifies low-dimensional subspaces of the representations that encode predefined sets of attributes. We find that these subspaces are smallest when they encode linguistic attributes. The subspaces also exhibit hierarchical structure present in the variables they encode and appear to be distributed across neurons. Ablation experiments reveal that BERT relies on subspaces with as few as 3 dimensions to make fine-grained part of speech distinctions when enforcing subject--verb agreement. Future work might explore richer geometric structure in word representations and its effect on model behavior.

\section*{Acknowledgments}
We would like to thank Jon Gauthier, Noga Zaslavsky, Peng Qian, Roger Levy, and John Hewitt for their helpful discussions and insightful comments, as well as the anonymous reviewers for the detailed feedback. Extra thanks to Jon and Roger for providing alpha access to Syntax Gym. This work was partially supported by a gift from NVIDIA under the NVAIL grant program.

\bibliography{anthology,custom}

\begin{thebibliography}{35}
\expandafter\ifx\csname natexlab\endcsname\relax\def\natexlab#1{#1}\fi

\bibitem[{Bau* et~al.(2019)Bau*, Belinkov*, Sajjad, Dalvi, Durrani, and
  Glass}]{BelinkovNeurons}
D.~Anthony Bau*, Yonatan Belinkov*, Hassan Sajjad, Fahim Dalvi, Nadir Durrani,
  and James Glass. 2019.
\newblock Identifying and controlling important neurons in neural machine
  translation.
\newblock In \emph{International Conference on Learning Representations
  (ICLR)}.

\bibitem[{Bau et~al.(2017)Bau, Zhou, Khosla, Oliva, and Torralba}]{NetDissect}
David Bau, Bolei Zhou, Aditya Khosla, Aude Oliva, and Antonio Torralba. 2017.
\newblock Network dissection: Quantifying interpretability of deep visual
  representations.
\newblock In \emph{Computer Vision and Pattern Recognition}.

\bibitem[{Belinkov and Glass(2019)}]{BelinkovSurvey}
Yonatan Belinkov and James Glass. 2019.
\newblock \href {https://doi.org/10.1162/tacl_a_00254} {{Analysis Methods in
  Neural Language Processing: A Survey}}.
\newblock \emph{Transactions of the Association for Computational Linguistics},
  7:49--72.

\bibitem[{Bolukbasi et~al.(2016)Bolukbasi, Chang, Zou, Saligrama, and
  Kalai}]{bolukbasi2016man}
Tolga Bolukbasi, Kai-Wei Chang, James~Y Zou, Venkatesh Saligrama, and Adam~T
  Kalai. 2016.
\newblock Man is to computer programmer as woman is to homemaker? debiasing
  word embeddings.
\newblock In \emph{Advances in neural information processing systems}, pages
  4349--4357.

\bibitem[{Coenen et~al.(2019)Coenen, Reif, Yuan, Kim, Pearce, Vi{\'e}gas, and
  Wattenberg}]{coenen2019visualizing}
Andy Coenen, Emily Reif, Ann Yuan, Been Kim, Adam Pearce, F.~Vi{\'e}gas, and
  M.~Wattenberg. 2019.
\newblock Visualizing and measuring the geometry of bert.
\newblock In \emph{NeurIPS}.

\bibitem[{Conneau et~al.(2018)Conneau, Kruszewski, Lample, Barrault, and
  Baroni}]{Conneau}
Alexis Conneau, German Kruszewski, Guillaume Lample, Lo{\"\i}c Barrault, and
  Marco Baroni. 2018.
\newblock \href {https://doi.org/10.18653/v1/P18-1198} {What you can cram into
  a single {\$}{\&}!{\#}* vector: Probing sentence embeddings for linguistic
  properties}.
\newblock In \emph{Proceedings of the 56th Annual Meeting of the Association
  for Computational Linguistics (Volume 1: Long Papers)}, pages 2126--2136,
  Melbourne, Australia. Association for Computational Linguistics.

\bibitem[{Dalvi et~al.(2019)Dalvi, Durrani, Sajjad, Belinkov, Bau, and
  Glass}]{GrainOfSand}
Fahim Dalvi, Nadir Durrani, Hassan Sajjad, Yonatan Belinkov, A.~Bau, and
  James~R. Glass. 2019.
\newblock What is one grain of sand in the desert? analyzing individual neurons
  in deep nlp models.
\newblock In \emph{AAAI}.

\bibitem[{Devlin et~al.(2019)Devlin, Chang, Lee, and Toutanova}]{BERT}
Jacob Devlin, Ming-Wei Chang, Kenton Lee, and Kristina Toutanova. 2019.
\newblock \href {https://doi.org/10.18653/v1/N19-1423} {{BERT}: Pre-training of
  deep bidirectional transformers for language understanding}.
\newblock In \emph{Proceedings of the 2019 Conference of the North {A}merican
  Chapter of the Association for Computational Linguistics: Human Language
  Technologies, Volume 1 (Long and Short Papers)}, pages 4171--4186,
  Minneapolis, Minnesota. Association for Computational Linguistics.

\bibitem[{Elazar et~al.(2021)Elazar, Ravfogel, Jacovi, and
  Goldberg}]{elazar2021amnesic}
Yanai Elazar, Shauli Ravfogel, Alon Jacovi, and Yoav Goldberg. 2021.
\newblock \href {https://doi.org/10.1162/tacl_a_00359} {{Amnesic Probing:
  Behavioral Explanation with Amnesic Counterfactuals}}.
\newblock \emph{Transactions of the Association for Computational Linguistics},
  9:160--175.

\bibitem[{Ethayarajh(2019)}]{ethayarajh2019contextual}
Kawin Ethayarajh. 2019.
\newblock \href {https://doi.org/10.18653/v1/D19-1006} {How contextual are
  contextualized word representations? {C}omparing the geometry of {BERT},
  {ELM}o, and {GPT}-2 embeddings}.
\newblock In \emph{Proceedings of the 2019 Conference on Empirical Methods in
  Natural Language Processing and the 9th International Joint Conference on
  Natural Language Processing (EMNLP-IJCNLP)}, pages 55--65, Hong Kong, China.
  Association for Computational Linguistics.

\bibitem[{Futrell et~al.(2019)Futrell, Wilcox, Morita, Qian, Ballesteros, and
  Levy}]{NLMSyntacticState}
Richard Futrell, Ethan Wilcox, Takashi Morita, Peng Qian, Miguel Ballesteros,
  and Roger Levy. 2019.
\newblock \href {https://doi.org/10.18653/v1/N19-1004} {Neural language models
  as psycholinguistic subjects: Representations of syntactic state}.
\newblock In \emph{Proceedings of the 2019 Conference of the North {A}merican
  Chapter of the Association for Computational Linguistics: Human Language
  Technologies, Volume 1 (Long and Short Papers)}, pages 32--42, Minneapolis,
  Minnesota. Association for Computational Linguistics.

\bibitem[{Gauthier et~al.(2020)Gauthier, Hu, Wilcox, Qian, and
  Levy}]{SyntaxGym}
Jon Gauthier, Jennifer Hu, Ethan Wilcox, Peng Qian, and Roger Levy. 2020.
\newblock \href {https://www.aclweb.org/anthology/2020.acl-demos.10}
  {{S}yntax{G}ym: An online platform for targeted evaluation of language
  models}.
\newblock In \emph{Proceedings of the 58th Annual Meeting of the Association
  for Computational Linguistics: System Demonstrations}, pages 70--76, Online.
  Association for Computational Linguistics.

\bibitem[{Hewitt and Liang(2019)}]{ControlTasks}
John Hewitt and Percy Liang. 2019.
\newblock Designing and interpreting probes with control tasks.
\newblock In \emph{Conference on Empirical Methods in Natural Language
  Processing}. Association for Computational Linguistics.

\bibitem[{Hewitt and Manning(2019)}]{StructuralProbe}
John Hewitt and Christopher~D. Manning. 2019.
\newblock \href {https://doi.org/10.18653/v1/N19-1419} {{A} structural probe
  for finding syntax in word representations}.
\newblock In \emph{Proceedings of the 2019 Conference of the North {A}merican
  Chapter of the Association for Computational Linguistics: Human Language
  Technologies, Volume 1 (Long and Short Papers)}, pages 4129--4138,
  Minneapolis, Minnesota. Association for Computational Linguistics.

\bibitem[{Kingma and Ba(2014)}]{Adam}
Diederik Kingma and Jimmy Ba. 2014.
\newblock Adam: A method for stochastic optimization.
\newblock In \emph{International Conference on Learning Representations}.

\bibitem[{Liu et~al.(2019)Liu, Gardner, Belinkov, Peters, and Smith}]{Liu}
Nelson~F. Liu, Matt Gardner, Yonatan Belinkov, Matthew~E. Peters, and Noah~A.
  Smith. 2019.
\newblock Linguistic knowledge and transferability of contextual
  representations.
\newblock In \emph{Proceedings of the Conference of the North American Chapter
  of the Association for Computational Linguistics: Human Language
  Technologies}.

\bibitem[{Marcus et~al.(1993)Marcus, Marcinkiewicz, and Santorini}]{PTB}
Mitchell~P. Marcus, Mary~Ann Marcinkiewicz, and Beatrice Santorini. 1993.
\newblock Building a large annotated corpus of english: The penn treebank.
\newblock \emph{Comput. Linguist.}, 19(2):313–330.

\bibitem[{Marvin and Linzen(2018)}]{marvin-linzen-2018-targeted}
Rebecca Marvin and Tal Linzen. 2018.
\newblock \href {https://doi.org/10.18653/v1/D18-1151} {Targeted syntactic
  evaluation of language models}.
\newblock In \emph{Proceedings of the 2018 Conference on Empirical Methods in
  Natural Language Processing}, pages 1192--1202, Brussels, Belgium.
  Association for Computational Linguistics.

\bibitem[{Mimno and Thompson(2017)}]{mimno2017strange}
David Mimno and Laure Thompson. 2017.
\newblock The strange geometry of skip-gram with negative sampling.
\newblock In \emph{Empirical Methods in Natural Language Processing}.

\bibitem[{Peters et~al.(2018)Peters, Neumann, Iyyer, Gardner, Clark, Lee, and
  Zettlemoyer}]{ELMo}
Matthew~E. Peters, Mark Neumann, Mohit Iyyer, Matt Gardner, Christopher Clark,
  Kenton Lee, and Luke Zettlemoyer. 2018.
\newblock \href {https://doi.org/10.18653/v1/N18-1202} {Deep contextualized
  word representations}.
\newblock In \emph{Proceedings of the 2018 Conference of the North {A}merican
  Chapter of the Association for Computational Linguistics: Human Language
  Technologies, Volume 1 (Long Papers)}, pages 2227--2237, New Orleans,
  Louisiana. Association for Computational Linguistics.

\bibitem[{Peters et~al.(2019)Peters, Ruder, and Smith}]{peters2019tune}
Matthew~E Peters, Sebastian Ruder, and Noah~A Smith. 2019.
\newblock To tune or not to tune? adapting pretrained representations to
  diverse tasks.
\newblock \emph{arXiv preprint arXiv:1903.05987}.

\bibitem[{Pimentel et~al.(2020{\natexlab{a}})Pimentel, Saphra, Williams, and
  Cotterell}]{pimentel2020pareto}
Tiago Pimentel, Naomi Saphra, Adina Williams, and Ryan Cotterell.
  2020{\natexlab{a}}.
\newblock \href {https://doi.org/10.18653/v1/2020.emnlp-main.254} {{P}areto
  probing: {T}rading off accuracy for complexity}.
\newblock In \emph{Proceedings of the 2020 Conference on Empirical Methods in
  Natural Language Processing (EMNLP)}, pages 3138--3153, Online. Association
  for Computational Linguistics.

\bibitem[{Pimentel et~al.(2020{\natexlab{b}})Pimentel, Valvoda, Hall~Maudslay,
  Zmigrod, Williams, and Cotterell}]{Pimentel}
Tiago Pimentel, Josef Valvoda, Rowan Hall~Maudslay, Ran Zmigrod, Adina
  Williams, and Ryan Cotterell. 2020{\natexlab{b}}.
\newblock \href {https://doi.org/10.18653/v1/2020.acl-main.420}
  {Information-theoretic probing for linguistic structure}.
\newblock In \emph{Proceedings of the 58th Annual Meeting of the Association
  for Computational Linguistics}, pages 4609--4622, Online. Association for
  Computational Linguistics.

\bibitem[{{Radford} et~al.(2017){Radford}, {Jozefowicz}, and
  {Sutskever}}]{SentimentNeuron}
Alec {Radford}, Rafal {Jozefowicz}, and Ilya {Sutskever}. 2017.
\newblock \href {http://arxiv.org/abs/1704.01444} {{Learning to Generate
  Reviews and Discovering Sentiment}}.
\newblock \emph{arXiv e-prints}, page arXiv:1704.01444.

\bibitem[{Ravfogel et~al.(2020)Ravfogel, Elazar, Gonen, Twiton, and
  Goldberg}]{INLP}
Shauli Ravfogel, Yanai Elazar, Hila Gonen, Michael Twiton, and Yoav Goldberg.
  2020.
\newblock \href {https://doi.org/10.18653/v1/2020.acl-main.647} {Null it out:
  Guarding protected attributes by iterative nullspace projection}.
\newblock In \emph{Proceedings of the 58th Annual Meeting of the Association
  for Computational Linguistics}, pages 7237--7256, Online. Association for
  Computational Linguistics.

\bibitem[{Ravichander et~al.(2021)Ravichander, Belinkov, and
  Hovy}]{ProbingProbing}
Abhilasha Ravichander, Yonatan Belinkov, and Eduard Hovy. 2021.
\newblock \href {https://aclanthology.org/2021.eacl-main.295} {Probing the
  probing paradigm: Does probing accuracy entail task relevance?}
\newblock In \emph{Proceedings of the 16th Conference of the European Chapter
  of the Association for Computational Linguistics: Main Volume}, pages
  3363--3377, Online. Association for Computational Linguistics.

\bibitem[{Saphra and Lopez(2019)}]{SaphraLopez}
Naomi Saphra and Adam Lopez. 2019.
\newblock \href {https://doi.org/10.18653/v1/N19-1329} {Understanding learning
  dynamics of language models with {SVCCA}}.
\newblock In \emph{Proceedings of the 2019 Conference of the North {A}merican
  Chapter of the Association for Computational Linguistics: Human Language
  Technologies, Volume 1 (Long and Short Papers)}, pages 3257--3267,
  Minneapolis, Minnesota. Association for Computational Linguistics.

\bibitem[{Shi et~al.(2016)Shi, Padhi, and Knight}]{shi2016does}
Xing Shi, Inkit Padhi, and Kevin Knight. 2016.
\newblock Does string-based neural mt learn source syntax?
\newblock In \emph{Proceedings of the 2016 Conference on Empirical Methods in
  Natural Language Processing}, pages 1526--1534.

\bibitem[{Tenney et~al.(2019{\natexlab{a}})Tenney, Das, and
  Pavlick}]{TenneyBertRelearns}
Ian Tenney, Dipanjan Das, and Ellie Pavlick. 2019{\natexlab{a}}.
\newblock \href {https://doi.org/10.18653/v1/P19-1452} {{BERT} rediscovers the
  classical {NLP} pipeline}.
\newblock In \emph{Proceedings of the 57th Annual Meeting of the Association
  for Computational Linguistics}, pages 4593--4601, Florence, Italy.
  Association for Computational Linguistics.

\bibitem[{Tenney et~al.(2019{\natexlab{b}})Tenney, Xia, Chen, Wang, Poliak,
  {Thomas McCoy}, Kim, {Van Durme}, Bowman, Das, and
  Pavlick}]{TenneyWhatDoYouLearn}
Ian Tenney, Patrick Xia, Berlin Chen, Alex Wang, Adam Poliak, R.~{Thomas
  McCoy}, Najoung Kim, Benjamin {Van Durme}, {Samuel R.} Bowman, Dipanjan Das,
  and Ellie Pavlick. 2019{\natexlab{b}}.
\newblock What do you learn from context? probing for sentence structure in
  contextualized word representations.
\newblock In \emph{Proceedings of the 7th International Conference on Learning
  Representations}.

\bibitem[{Vargas and Cotterell(2020)}]{vargas2020exploring}
Francisco Vargas and Ryan Cotterell. 2020.
\newblock \href {https://doi.org/10.18653/v1/2020.emnlp-main.232} {Exploring
  the linear subspace hypothesis in gender bias mitigation}.
\newblock In \emph{Proceedings of the 2020 Conference on Empirical Methods in
  Natural Language Processing (EMNLP)}, pages 2902--2913, Online. Association
  for Computational Linguistics.

\bibitem[{Voita et~al.(2019)Voita, Sennrich, and Titov}]{VoitaTransformers}
Elena Voita, Rico Sennrich, and Ivan Titov. 2019.
\newblock \href {https://doi.org/10.18653/v1/D19-1448} {The bottom-up evolution
  of representations in the transformer: A study with machine translation and
  language modeling objectives}.
\newblock In \emph{Proceedings of the 2019 Conference on Empirical Methods in
  Natural Language Processing and the 9th International Joint Conference on
  Natural Language Processing (EMNLP-IJCNLP)}, pages 4396--4406, Hong Kong,
  China. Association for Computational Linguistics.

\bibitem[{Voita and Titov(2020)}]{Voita}
Elena Voita and Ivan Titov. 2020.
\newblock \href {https://doi.org/10.18653/v1/2020.emnlp-main.14}
  {Information-theoretic probing with minimum description length}.
\newblock In \emph{Proceedings of the 2020 Conference on Empirical Methods in
  Natural Language Processing (EMNLP)}, pages 183--196, Online. Association for
  Computational Linguistics.

\bibitem[{Wolf et~al.(2020)Wolf, Debut, Sanh, Chaumond, Delangue, Moi, Cistac,
  Rault, Louf, Funtowicz, Davison, Shleifer, von Platen, Ma, Jernite, Plu, Xu,
  Scao, Gugger, Drame, Lhoest, and Rush}]{HuggingFace}
Thomas Wolf, Lysandre Debut, Victor Sanh, Julien Chaumond, Clement Delangue,
  Anthony Moi, Pierric Cistac, Tim Rault, Rémi Louf, Morgan Funtowicz, Joe
  Davison, Sam Shleifer, Patrick von Platen, Clara Ma, Yacine Jernite, Julien
  Plu, Canwen Xu, Teven~Le Scao, Sylvain Gugger, Mariama Drame, Quentin Lhoest,
  and Alexander~M. Rush. 2020.
\newblock \href {https://www.aclweb.org/anthology/2020.emnlp-demos.6}
  {Transformers: State-of-the-art natural language processing}.
\newblock In \emph{Proceedings of the 2020 Conference on Empirical Methods in
  Natural Language Processing: System Demonstrations}, pages 38--45, Online.
  Association for Computational Linguistics.

\bibitem[{Zhang and Bowman(2018)}]{TranslationProbing}
Kelly Zhang and Samuel Bowman. 2018.
\newblock \href {https://doi.org/10.18653/v1/W18-5448} {Language modeling
  teaches you more than translation does: Lessons learned through auxiliary
  syntactic task analysis}.
\newblock In \emph{Proceedings of the 2018 {EMNLP} Workshop {B}lackbox{NLP}:
  Analyzing and Interpreting Neural Networks for {NLP}}, pages 359--361,
  Brussels, Belgium. Association for Computational Linguistics.

\end{thebibliography}
\bibliographystyle{acl_natbib}

\appendix

\section{Linear Probes}
\label{app:linear-probes}

\begin{table*}
    \centering
    \footnotesize
    \begin{tabular}{ccllllll}
    \toprule
    \multicolumn{2}{c}{\multirow{2}{*}{}} & \multicolumn{2}{c}{\textbf{POS}} & \multicolumn{2}{c}{\textbf{DLP}}  & \multicolumn{2}{c}{\textbf{DEP}} \\ %
    \multicolumn{2}{c}{} & \multicolumn{1}{c}{Real} & \multicolumn{1}{c}{Control} & \multicolumn{1}{c}{Real} & \multicolumn{1}{c}{Control} & \multicolumn{1}{c}{Real} & \multicolumn{1}{c}{Control} \\ \midrule %
    \multicolumn{1}{c}{\multirow{3}{*}{\textbf{ELMo}}} & 0  & .93 / .93 / 6 & .75 / .77 / 256 & .87 / .85 / 5 & .21 / .21 / 512 & .33 / .35 / 11 & .66 / .68 / 13 \\
    \multicolumn{1}{c}{} & 1  & .97 / .94 / 5  & .70 / .72 / 256 & .96 / .91 / 3 & .23 / .23 / 512 & .85 / .80 / 13 & .81 / .77 / 17 \\ 
    \multicolumn{1}{c}{} & 2  & .97 / .93 / 6 & .65 / .65 / 256 & .96 / .91 / 4 & .21 / .21 / 256 & .80 / .72 / 21  & .76 / .69 / 23 \\ \midrule %
    \multicolumn{1}{c}{\multirow{5}{*}{\textbf{BERT}}} & 0  & .83 / .81 / 4 & .74 / .75 / 26 & .81 / .79 / 6 & .22 / .22 / 256 & .46 / .50 / 13 & .75 / .75 / 7 \\
    \multicolumn{1}{c}{} & 1  & .89 / .85 / 5 & .73 / .73 / 128 & .88 / .86 / 6  & .22 / .22 / 256 & .59 / .63 / 17 & .78 / .76 / 9 \\ 
    \multicolumn{1}{c}{} & 4  & .90 / .88 / 6 & .69 / .69 / 128 & .91 / .89 / 5 & .21 / .21 / 256 & .74 / .74 / 13 & .78 / .75 / 10 \\ 
    \multicolumn{1}{c}{} & 8  & .91 / .87 / 7 & .60 / .60 / 256 & .92 / .89 / 5 & .19 / .18 / 64 & .80 / .75 / 13 & .70 / .67 / 12 \\ 
    \multicolumn{1}{c}{} & 12 & .88 / .85 / 10 & .53 / .54 / 256 & .89 / .86 / 6 & .18 / .17 / 24 & .73 / .72 / 20 & .65 / .64 / 11 \\ \bottomrule %
    \end{tabular}
    \caption{
     Linear probe accuracy for each representation and task. Each entry reports three numbers:
     \textbf{accuracy} of a linear probe trained on the full representation / \textbf{accuracy} of
     a linear probe trained on projected representations / \textbf{dimensionality} of the
     projection. Dimensionalities are the optimal values of $d$ found using the MLP probe in
     \Cref{sec:sweep} and shown previously in \Cref{tab:acc-vs-dim}.
    }
    \label{tab:linear-acc-vs-dim}
\end{table*}

Our MLP probes from \Cref{sec:sweep} learn to predict linguistic features in low-dimensional linear subspaces of the representations. However, this does not mean the variables are \emph{linearly encoded} in the subspaces. It is possible that the MLP relies on nonlinear information to make its predictions.

We investigate by swapping the MLP probe from \Cref{sec:sweep} with a linear softmax classifier. For each task, we train two probes: one with projection rank $D$ and one with the same projection rank as the optimal MLP probes presented in \Cref{tab:acc-vs-dim}. If the linear probes obtain lower accuracy than the MLP probes, we can conclude the subspace nonlinearly encodes the task.

\Cref{tab:linear-acc-vs-dim} shows that this is indeed the case. The linear probes generally fall short of the MLP probes across tasks, even when the classifier is not rank constrained. We observe, however, that the discrepancies are not large, suggesting that most of the task-relevant variables are  linear.

\section{Probing Untrained Representations}
\label{sec:random}

Do \emph{untrained} word representations also contain low-dimensional subspaces that encode linguistic features? 
We repeat the experiment from \Cref{sec:sweep}, but this time obtain representations from a randomly initialized, untrained BERT model. \Cref{tab:random-bert} shows the results.
All probes top off at considerably lower accuracies than they do when applied to trained representations (\Cref{tab:acc-vs-dim}), suggesting the linguistic information is more scarcely encoded. This discrepancy is smaller for the real POS and DEP tasks, but for POS the probe still requires a higher-dimensional subspace than when it is applied to trained representations (e.g., 14 dimensions instead of 5 for layer 1).

\begin{table*}[t]
    \centering
    \footnotesize
    \begin{tabular}{ccllllll}
    \toprule
    \multicolumn{2}{c}{\multirow{2}{*}{}} & \multicolumn{2}{c}{\textbf{POS}} & \multicolumn{2}{c}{\textbf{DLP}}  & \multicolumn{2}{c}{\textbf{DEP}} \\ %
    \multicolumn{2}{c}{} & \multicolumn{1}{c}{Real} & \multicolumn{1}{c}{Control} & \multicolumn{1}{c}{Real} & \multicolumn{1}{c}{Control} & \multicolumn{1}{c}{Real} & \multicolumn{1}{c}{Control} \\ \midrule %
    \multicolumn{1}{c}{\multirow{5}{*}{\textbf{Untrained BERT}}} & 0  & .74 / 14 & .09 / 1 & .13 / 4 & .11 / 1 & .52 / 18 & .37 / 2 \\
    \multicolumn{1}{c}{} & 1  & .72 / 14 & .10 / 2 & .12 / 3  & .10 / 1 & .50 / 15 & .30 / 1 \\ 
    \multicolumn{1}{c}{} & 4  & .73 / 13 & .11 / 6 & .12 / 3 & .11 / 1 & .48 / 14 & .34 / 1 \\ 
    \multicolumn{1}{c}{} & 8  & .68 / 9 & .14 / 1 & .13 / 6 & .11 / 1 & .43 / 12 & .34 / 1 \\ 
    \multicolumn{1}{c}{} & 12 & .69 / 12 & .11 / 10 & .12 / 6 & .07 / 1 & .36 / 6 & .37 / 1 \\ \bottomrule %
    \end{tabular}
    \caption{
    Probe accuracy and dimensionality for randomly initialized BERT representations. Compared to \Cref{tab:acc-vs-dim}, the low accuracies suggest the linguistic information is not well encoded at all in the representations, with the exceptions being real POS and real DEP.
    }
    \label{tab:random-bert}
\end{table*}

\section{INLP Ablations}
\label{app:inlp}

\begin{table}
    \centering
    \footnotesize
    \begin{tabular}{lccc}
    \toprule
    \multirow{2}{*}{\begin{tabular}[c]{@{}l@{}}Marginal\\Probability\end{tabular}} & %
    & Ablated &
    \\ \cmidrule{2-4}
    & Nothing & Verbspace & Nounspace \\ \cmidrule{1-4}
    Nouns & .85  & .82 & \textbf{.74} \\ \cmidrule{1-4}
    Verbs & .54  & \textbf{.38} & .53  \\ \bottomrule
    \end{tabular}
    \caption{
        Same as \Cref{tab:ablation}, but now Nounspace and Verbspace are computed using INLP.
    }
    \label{tab:ablation-inlp}
\end{table}

\begin{figure}
    \vspace{-0.5em}
    \centering
    \includegraphics[width=.5\textwidth]{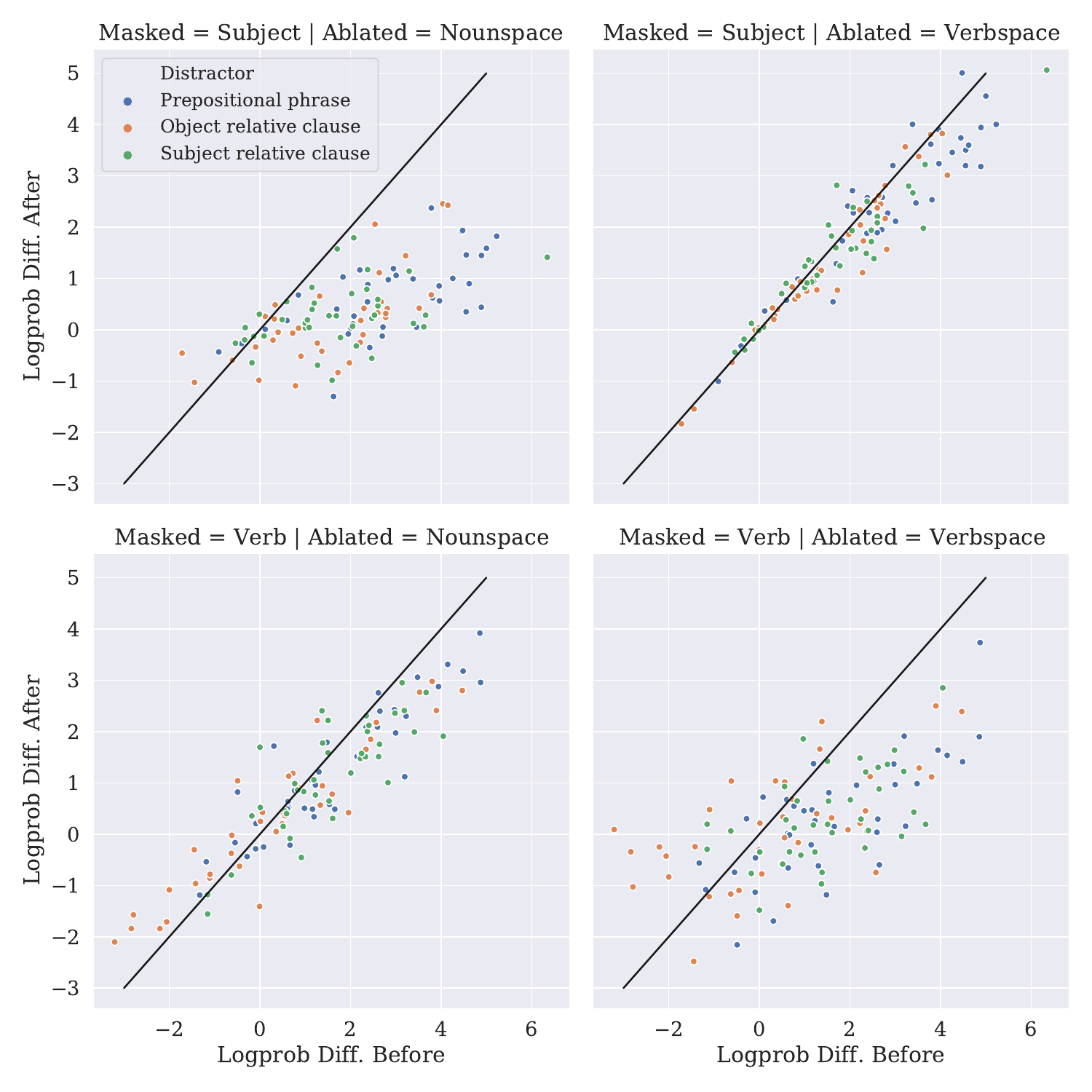}
    \caption{
        Same as \Cref{fig:diffs}, but now nounspace and verbspace are computed using INLP.
    }
    \label{fig:diffs-inlp}
    \vspace{-1em}
\end{figure}

In \Cref{sec:ablation}, we present a method for removing the minimal linear subspace that encodes a task. INLP \cite{INLP} is a method for \emph{maximally} removing the linear information that encodes a task. For comparison, we present the results of our ablation experiment when nounspace and verbspace are computed using INLP instead of our method.

We begin with a brief overview of INLP. Given representations $R$ and a task $f$ with $k$ labels, INLP removes all information from $R$ that is linearly predictive of $f$. It does this by iteratively constructing a projection $P$ such that no linear classifier can predict $f$ from $PR$. At iteration $i$, INLP trains a linear classifier $W^{(i)} \in \mathbb{R}^{n \times k}$ to predict $F$ from the projected representations $P^{(i-1)}R$. It then computes
$$P^{(i)} = \textsc{Proj}\big\{\textsc{Null}(W^{(1)}) \cap \dots \cap \textsc{Null}(W^{(i)})\}$$
where \textsc{Proj} is the operation that computes a projection onto a subspace, and \textsc{Null} is an operation computing the nullspace of a linear transformation. The algorithm terminates when the optimal $W^{(i)}$ always predicts the majority class.

\Cref{tab:ablation-inlp} and \Cref{fig:diffs-inlp} replicate \Cref{tab:ablation} and \Cref{fig:diffs} from \Cref{sec:ablation} using the verb- and nounspaces computed with INLP. We see the same trends discussed in \Cref{ssec:ablation-results}, but more pronounced. One key difference is that the INLP ablations cause substantial but isolated change to how BERT assigns probability mass to different parts of speech. For example, ablating the maximal verbspace impairs BERT's ability to predict verbs in the verb slot (probability .54 before vs. .38 after), but not its ability to predict nouns in the subject slot (probability .85 before vs. .82 after). This suggests that INLP is more destructive than our method: by removing so much linear information from the representations, it not only ablates BERT's ability to distinguish verb types but also its ability to distinguish verbs from other parts of speech.

\end{document}